\documentclass[10pt,twocolumn,letterpaper]{article}

\usepackage{cvpr}      

\usepackage{graphicx}
\usepackage{multirow}

\newif\iffinalversion
\makeatletter
\@ifpackagewith{cvpr}{review}{\finalversionfalse}{\finalversiontrue}
\makeatother

\definecolor{cvprblue}{rgb}{0.21,0.49,0.74}
\usepackage[pagebackref,breaklinks,colorlinks,allcolors=cvprblue]{hyperref}

\def\paperID{29} 
\def\confName{CVPR}
\def\confYear{2026}

\title{Better Rigs, Not Bigger Networks: A Body Model Ablation for Gaussian Avatars}

\author{Derek Austin\\
{\tt\small dcaustin33@gmai1.com}
}

\begin{document}
\maketitle

\begin{abstract}
Recent 3D Gaussian splatting methods built atop SMPL achieve remarkable visual fidelity while continually increasing the complexity of the overall training architecture. 
We demonstrate that much of this complexity is unnecessary: by replacing SMPL with the Momentum Human Rig (MHR), estimated via SAM-3D-Body, a minimal pipeline with no learned deformations or pose-dependent corrections achieves the highest reported PSNR and competitive or superior LPIPS and SSIM on PeopleSnapshot and ZJU-MoCap. 
To disentangle pose estimation quality from body model representational capacity, we perform two controlled ablations: translating SAM-3D-Body meshes to SMPL-X, and translating the original dataset's SMPL poses into MHR both retrained under identical conditions.
These ablations confirm that body model expressiveness has been a primary bottleneck in avatar reconstruction, with both mesh representational capacity and pose estimation quality contributing meaningfully to the full pipeline's gains.

\end{abstract}

\iffinalversion
  \begingroup
  \renewcommand{\thefootnote}{}
  \footnotetext{Code: \url{https://github.com/dcaustin33/better_rigs_not_bigger_networks}}
  \endgroup
\fi

\section{Introduction}
\label{sec:intro}
Realistic human avatars have a wide range of impactful use cases, especially in the fields of gaming, simulation and AR/VR.
Recently, many works have made substantial leaps and achieved high visual quality, but each new method introduces additional architectural complexity to handle failure modes.
Many focus on deformation adjustments, such as custom-built MLPs to compensate for clothing and pose-dependent appearance shifts~\cite{3DGSAvatar, rmavatar, GaussianAvatar}.
What these methods share, however, is a dependence on SMPL~\cite{SMPL} as the body model.

We show that upgrading the upstream body model reduces the need for this downstream complexity. 
Specifically, replacing SMPL with the Momentum Human Rig~\cite{MHR} (127 joints and 18.4k vertices compared to 24 joints and 6,890 vertices in SMPL) estimated via the SAM-3D-Body model~\cite{sam3dbody} achieves the highest reported PSNR values and competitive SSIM and LPIPS values in a minimal Gaussian splatting pipeline.
To disentangle pose estimation quality from body model representational capacity, we translate our SAM-3D-Body meshes back to SMPL-X (due to the higher level of detail than SMPL) using the fitting tools of SAM-3D-Body~\cite{sam3dbody} and retrain under identical conditions. 
We also isolate the contribution that pose quality alone gives us by translating from the given SMPL poses into MHR.
Our ablations reveal that both pose estimation quality and mesh representational capacity contribute meaningfully to reconstruction fidelity. 
Crucially, MHR provides gains along both axes simultaneously.
Its richer skeleton and recent pose estimation model yields more accurate pose fitting while its denser mesh better supports Gaussian attachment, explaining why the full pipeline's improvement exceeds either factor in isolation.

\smallskip
\noindent Our contributions are as follows:
\begin{itemize}
  \item A minimal Gaussian avatar pipeline that, by substituting MHR for SMPL and removing all learned deformation modules, achieves the highest reported PSNR on both PeopleSnapshot and ZJU-MoCap.
  \item A controlled ablation that translates poses between body models under identical training conditions, disentangling the contributions of pose estimation quality and mesh representational capacity.
\end{itemize}

\begin{table*}[t]
\centering
\caption{Quantitative comparison on the PeopleSnapshot dataset. We report PSNR$\uparrow$, SSIM$\uparrow$, and LPIPS$\downarrow$ for novel pose synthesis. Best results are in \textbf{bold}, second best are \underline{underlined}. All numbers listed are provided by authors.}
\label{tab:peoplesnapshot}
\resizebox{\textwidth}{!}{%
\begin{tabular}{l ccc ccc ccc ccc ccc}
\toprule
& \multicolumn{3}{c}{Male-3-casual} & \multicolumn{3}{c}{Male-4-casual} & \multicolumn{3}{c}{Female-3-casual} & \multicolumn{3}{c}{Female-4-casual} & \multicolumn{3}{c}{Avg} \\
\cmidrule(lr){2-4} \cmidrule(lr){5-7} \cmidrule(lr){8-10} \cmidrule(lr){11-13} \cmidrule(lr){14-16}
Method & PSNR$\uparrow$ & SSIM$\uparrow$ & LPIPS$\downarrow$
       & PSNR$\uparrow$ & SSIM$\uparrow$ & LPIPS$\downarrow$
       & PSNR$\uparrow$ & SSIM$\uparrow$ & LPIPS$\downarrow$
       & PSNR$\uparrow$ & SSIM$\uparrow$ & LPIPS$\downarrow$
       & PSNR$\uparrow$ & SSIM$\uparrow$ & LPIPS$\downarrow$ \\
\midrule
Anim-NeRF~\cite{AnimNerf}
  & 29.37 & 0.9703 & 0.0168
  & 28.37 & 0.9605 & 0.0268
  & 28.91 & 0.9743 & \underline{0.0215}
  & 28.90 & 0.9678 & 0.0174
  & 28.89 & 0.9682 & 0.0206 \\
InstantAvatar~\cite{InstantAvatar}
  & 30.91 & 0.977 & 0.022
  & 29.77 & \underline{0.980} & 0.025
  & 29.73 & 0.975 & 0.025
  & 30.92 & 0.977 & 0.021
  & 30.33 & 0.9773 & 0.0233 \\
GaussianAvatar~\cite{GaussianAvatar}
  & 30.98 & \underline{0.9790} & \textbf{0.0145}
  & 28.78 & 0.9755 & \textbf{0.0228}
  & 29.55 & 0.9762 & 0.0225
  & 30.84 & 0.9771 & \textbf{0.0140}
  & 30.04 & 0.9769 & \textbf{0.0185} \\
SplattingAvatar~\cite{SplattingAvatar}
  & 33.01 & \textbf{0.982} & 0.020
  & 30.99 & \textbf{0.982} & 0.029
  & 30.81 & \underline{0.978} & 0.028
  & 32.57 & \textbf{0.981} & 0.018
  & 31.84 & \textbf{0.9808} & 0.0238 \\
3DGS-Avatar~\cite{3DGSAvatar}
  & 34.28 & 0.9724 & \underline{0.0149}
  & 30.22 & 0.9653 & \underline{0.0231}
  & 30.57 & 0.9581 & \textbf{0.0209}
  & 33.16 & 0.9678 & \underline{0.0157}
  & 32.06 & 0.9659 & \underline{0.0186} \\
\midrule
Ours (SMPL-X)
  & \underline{34.65} & 0.9679 & 0.0281
  & \underline{32.32} & 0.9710 & 0.0343
  & \underline{35.04} & 0.9724 & 0.0333
  & \underline{34.91} & 0.9726 & 0.0235
  & \underline{34.23} & 0.9710 & 0.0298 \\
Ours (MHR)
  & \textbf{36.94} & 0.9776 & 0.0187
  & \textbf{34.20} & 0.9778 & 0.0265
  & \textbf{37.56} & \textbf{0.9787} & 0.0267
  & \textbf{37.01} & \underline{0.9801} & 0.0164
  & \textbf{36.43} & \underline{0.9786} & 0.0221 \\
\bottomrule
\end{tabular}%
}
\end{table*}

\section{Related Work}
\label{sec:related_work}
\subsection{Animatable Avatar Methods}
Animatable avatar reconstruction from monocular video has progressed rapidly from NeRF-based to Gaussian-based methods. 
HumanNeRF~\cite{humanNerf} established the paradigm of learning a canonical T-pose neural radiance field~\cite{nerf} paired with a skeleton-driven motion field, while also establishing evaluation protocols most commonly used today. 
InstantAvatar~\cite{InstantAvatar} combined hash-based NeRF acceleration with Fast-SNARF articulation, reducing training to under a minute and increasing rendering speeds to 15 FPS+.

The shift to Gaussian splatting allowed for faster rendering times along with a host of optimizations to account for persistent failure modes.
3DGS-Avatar~\cite{3DGSAvatar} uses a deformation module in concert with a multi-level hash grid to model pose-dependent effects in concert with an `As-Isometric-As-Possible` loss to ensure geometric consistency of Gaussian offsets from the canonical pose to the rendered pose.
Human Gaussian Splats~\cite{Kocabas2024HUGS} utilizes a feature triplane and optimizes LBS weights during training to account for clothing and human deformation during training.
GaussianAvatar~\cite{GaussianAvatar} uses a pose encoder with an optimizable feature tensor per Gaussian to predict pose-dependent offsets, color and scale.

Two methods adopt a simple strategy close to ours by embedding Gaussians directly on mesh triangles.
SplattingAvatar~\cite{SplattingAvatar} optimizes Gaussians directly on the given mesh while introducing a triangle-walking scheme to continually update the face each Gaussian is assigned to. 
GaussianAvatars~\cite{gaussianAvatars} rigs Gaussians to FLAME~\cite{FLAME} face mesh triangles without any deformation MLP, demonstrating that triangle-embedded Gaussians can produce high-quality results when the underlying mesh is sufficiently expressive.
Our method is most architecturally similar to GaussianAvatars, extending a triangle-embedded Gaussian approach from heads to full bodies. 

All of the above methods share a common dependency on SMPL as the upstream body model, and address its limitations exclusively through downstream architectural additions.
We instead ask whether upgrading the body model itself can eliminate the need for this complexity.

\subsection{Parametric Body Models}
SMPL~\cite{SMPL} has been the dominant body model in avatar reconstruction.
Its 24-joint skeleton with 6,890 vertices provides a compact, well-tooled representation, but its limited joint set lacks twist degrees of freedom in the limbs, producing well-known candy-wrapper artifacts under forearm pronation and volume loss at deeply flexed joints.
SMPL-X~\cite{SMPL-X:2019} extends SMPL with articulated hands via MANO~\cite{MANO:SIGGRAPHASIA:2017} and facial detail via FLAME~\cite{FLAME}, increasing the model to 54 joints and 10,475 vertices, but retains the same LBS formulation and joint-from-surface derivation as SMPL.

Momentum Human Rig offers a substantially richer mesh representation with 127 joints, 18,439 vertices, non-linear pose corrections, decoupled skeleton-mesh architectures and a partitioned identity space \cite{MHR}.
Its twist joints directly address the candy-wrapper artifacts that SMPL-based pipelines must learn to compensate for, and its denser mesh provides finer-grained Gaussian attachment.
In order to help estimate MHR parameters, SAM-3D-Body was also released which enables fitting of a mesh from a single image \cite{sam3dbody}.
Despite MHR's representational advantages, no prior avatar method, to our knowledge, has explored replacing SMPL with a higher-capacity body model.

\begin{table*}[t]
\centering
\caption{Quantitative comparison on the ZJU-MoCap dataset. We report PSNR$\uparrow$, SSIM$\uparrow$, and LPIPS$\downarrow$ for novel view synthesis. Best results are in \textbf{bold}, second best are \underline{underlined}. Fewer baselines report ZJU-MoCap numbers under the single-camera training protocol. All numbers listed are provided by authors. }
\label{tab:zjumocap}
\resizebox{\textwidth}{!}{%
\begin{tabular}{l ccc ccc ccc ccc ccc ccc ccc}
\toprule
& \multicolumn{3}{c}{377} & \multicolumn{3}{c}{386} & \multicolumn{3}{c}{387} & \multicolumn{3}{c}{392} & \multicolumn{3}{c}{393} & \multicolumn{3}{c}{394} & \multicolumn{3}{c}{Avg} \\
\cmidrule(lr){2-4} \cmidrule(lr){5-7} \cmidrule(lr){8-10} \cmidrule(lr){11-13} \cmidrule(lr){14-16} \cmidrule(lr){17-19} \cmidrule(lr){20-22}
Method & PSNR & SSIM & LPIPS
       & PSNR & SSIM & LPIPS
       & PSNR & SSIM & LPIPS
       & PSNR & SSIM & LPIPS
       & PSNR & SSIM & LPIPS
       & PSNR & SSIM & LPIPS
       & PSNR & SSIM & LPIPS \\
\midrule
GaussianAvatar~\cite{GaussianAvatar}
  & 24.86 & 0.944 & 0.063
  & 27.10 & 0.923 & 0.074
  & 25.63 & 0.947 & \underline{0.043}
  & 26.18 & 0.929 & 0.088
  & 23.90 & 0.919 & 0.099
  & 26.11 & 0.925 & 0.084
  & 25.63 & 0.9312 & 0.0752 \\
SplattingAvatar~\cite{SplattingAvatar}
  & 32.24 & 0.977 & 0.028
  & 30.31 & 0.953 & 0.073
  & \textbf{30.69} & \textbf{0.967} & 0.045
  & \textbf{33.41} & \textbf{0.975} & 0.040
  & \underline{30.02} & \underline{0.962} & 0.047
  & \underline{32.36} & \underline{0.965} & 0.044
  & 31.51 & 0.9665 & 0.0462 \\
3DGS-Avatar~\cite{3DGSAvatar}
  & 30.96 & 0.9778 & \textbf{0.0198}
  & 33.94 & \textbf{0.9784} & \textbf{0.0247}
  & 28.40 & \underline{0.9656} & \textbf{0.0330}
  & 32.10 & \underline{0.9739} & \textbf{0.0292}
  & 29.30 & \textbf{0.9645} & \textbf{0.0340}
  & 30.74 & \textbf{0.9662} & \textbf{0.0310}
  & 30.91 & \textbf{0.9711} & \textbf{0.0286} \\
\midrule
Ours (SMPL-X)
  & \underline{33.51} & \underline{0.9799} & 0.0241
  & \underline{35.43} & \underline{0.9722} & \underline{0.0323}
  & 29.72 & 0.9557 & 0.0492
  & 32.43 & 0.9635 & 0.0469
  & 29.63 & 0.9515 & 0.0567
  & 32.04 & 0.9599 & 0.0456
  & \underline{32.13} & 0.9638 & 0.0425 \\
Ours (MHR)
  & \textbf{34.26} & \textbf{0.9821} & \underline{0.0232}
  & \textbf{35.50} & 0.9716 & 0.0340
  & \underline{29.81} & 0.9603 & 0.0434
  & \underline{33.00} & 0.9687 & \underline{0.0395}
  & \textbf{30.21} & 0.9596 & \underline{0.0435}
  & \textbf{32.46} & 0.9644 & \underline{0.0388}
  & \textbf{32.54} & \underline{0.9678} & \underline{0.0371} \\
\bottomrule
\end{tabular}%
}
\end{table*}

\section{Method}
\label{sec:method}
\subsection{Overview}
Similar to GaussianAvatars \cite{gaussianAvatars}, we embed Gaussians directly on mesh triangles.
We start by assigning 30,000 Gaussians to our mesh with at least 1 Gaussian to each triangle followed by a random assignment for the remaining.
Each Gaussian stores its properties (position, rotation, scale, RGB color) in the local coordinate frame of the parent triangle.
We do not use any learnable feature tensor or spherical harmonics in order to represent color.
Per-triangle transforms, constructed from the posed vertices, rigidly map each Gaussian's local parameters to world space without any learned deformation.

\smallskip
\noindent A summary of the forward pass:
\begin{enumerate}
  \item Our body model maps pose $\theta$ and shape $\beta$ to posed vertices via LBS and pose corrections offered in MHR
  \item Per-triangle transforms (scaling and position) are then constructed from the posed vertices transforming the Gaussian local parameters to world space via each individual triangle's transforms
  \item Gsplat \cite{gsplat} is then used as a differentiable rasterizer to render each image
\end{enumerate}

\subsection{Training Details}
Our loss is constituted of an L1, SSIM and LPIPS reconstruction loss and small mask regularization, i.e. penalizing Gaussians rendered outside the given segmentation mask.
All code and hyperparameters are available in the supplementary materials.
We explicitly do not use any spherical harmonics, deformation MLPs, feature triplanes or specialized losses in order to isolate mesh types as the main source of quality.

We run each experiment for 50k iterations with the Adam optimizer \cite{adam} with an exponential decay schedule to 10\% of the initial learning rate. 
We start each experiment with 30k initial Gaussians and follow a typical split, prune and densify schedule every 500 iterations.
Once the total number of Gaussians has exceeded 100k we do not allow any further splitting or densifying.
The only variable we change across all of our experiments is the mesh representation.

\subsection{Body Mesh}
SAM-3D-Body estimates MHR parameters that we use to train MHR-specific models.
Those MHR parameters are then used to fit to SMPL-X via iterative optimization through tools provided in the SAM-3D-Body repository ~\cite{sam3dbody}.
We choose SMPL-X rather than SMPL as the ablation target due to it being the highest-capacity model in the SMPL family. 
SMPL-X's articulated hands and expressive face mesh ensure that any remaining quality gap cannot be attributed to SMPL simply lacking geometry in those regions.
All code is made available in the supplementary materials and our dataset will be made available to the community.

\section{Experiments}
\label{sec:experiments}
\subsection{Setup}

\textbf{Datasets.}
We evaluate on PeopleSnapshot~\cite{people_snapshot} (4 subjects: male-3-casual,
male-4-casual, female-3-casual, female-4-casual; monocular video, 540$\times$540)
and ZJU-MoCap~\cite{zju_mocap} (6 subjects: 377, 386, 387, 392, 393, 394;
multi-view capture, 512$\times$512). For PeopleSnapshot we adopt the temporal
train/test split of Anim-NeRF~\cite{AnimNerf}; for ZJU-MoCap we train on a
single camera (camera 1) and evaluate on held-out cameras following 3D-Gaussian-Avatars ~\cite{3DGSAvatar}.

\textbf{Metrics.}
We report PSNR, SSIM, and LPIPS (VGG backbone) to match published baselines. PSNR measures per-pixel similarity whereas SSIM and LPIPS provide structural and textural quality.

\subsection{Main Results}
Our pipeline achieves the highest average PSNR on both benchmarks while remaining competitive if not exceeding the state of the art on SSIM and LPIPS.
The PSNR advantage is notable because PSNR penalizes per-pixel error without spatial tolerance, making it the metric least amenable to compensation by learned texture and appearance corrections. 
Leading PSNR scores indicate that MHR's posed meshes provide a stable, faithful representation allowing the optimization to place Gaussians closer to the true surface than SMPL-based alternatives, producing geometrically more accurate renderings before any learned correction has a chance to act.

The narrower margins on SSIM and LPIPS are expected as both metrics are more sensitive to perceptual high-frequency texture quality, precisely the phenomena that color MLPs and pose-conditioned decoders in prior methods target. 
We deliberately disable these components to isolate the body model's contribution and provide a floor for MHR's performance. 
The remaining gap suggests that lightweight learned corrections on top of MHR would likely be useful, a direction we leave to future work.

One possible explanation of our results is that the deformation MLPs that many competing methods use serve as a stand-in for the non-linear pose correctives that MHR employs.

\subsection{Qualitative Comparison}
\begin{figure}[htbp]
    \centering
    \includegraphics[width=0.5\textwidth]{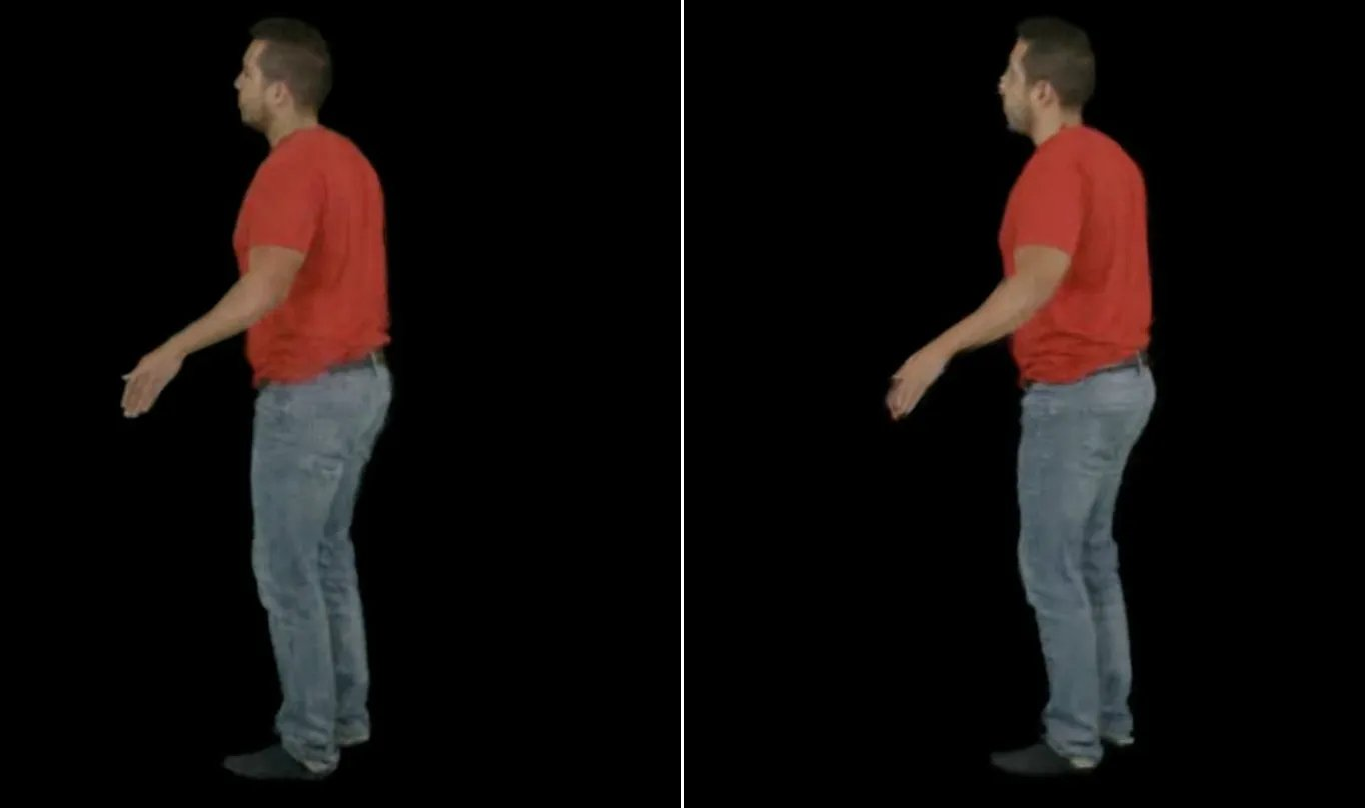}
    \caption{Rendering of Male-3-casual from the People Snapshot dataset with MHR mesh on the left and SMPL-X rendering on the right.}
    \label{fig:mhr_vs_smplx}
\end{figure}
We find that MHR provides far higher quality detail on tough-to-model body locations like hands, faces and elbows than even SMPL-X.
As shown in Figure \ref{fig:mhr_vs_smplx}, our model that utilizes MHR is far more visually realistic in hands and face especially as compared to SMPL-X.
We hypothesize that due to the more faithful and finer detail mesh that MHR provides, the optimization process is more stable, allowing the Gaussians to faithfully model hands and facial shape.

\begin{table}[t]
\centering
\caption{Impact of pose fitting and body model on rendering quality (PeopleSnapshot). Replacing Anim-NeRF's SMPL fits with SAM-3D-Body's MHR fits yields large gains (rows 1\,vs.\,3), and retaining the native MHR representation rather than converting to SMPL-X provides further improvement (rows 2\,vs.\,3). Best results are in \textbf{bold}, second best are \underline{underlined}.}
\label{tab:pose_quality}
\resizebox{\columnwidth}{!}{%
\begin{tabular}{ll l ccc}
\toprule
& Pose Fitting & Rendering Model & PSNR\,$\uparrow$ & SSIM\,$\uparrow$ & LPIPS\,$\downarrow$ \\
\midrule
\multirow{3}{*}{male-3-casual}
  & Anim-NeRF (SMPL)    & MHR    & 31.65              & 0.9538              & 0.0349              \\
  & SAM-3D-Body (MHR)  & SMPL-X & \underline{34.65}  & \underline{0.9679}  & \underline{0.0281}  \\
  & SAM-3D-Body (MHR)  & MHR    & \textbf{36.94}     & \textbf{0.9776}     & \textbf{0.0187}     \\
\midrule
\multirow{3}{*}{female-3-casual}
  & Anim-NeRF (SMPL)    & MHR    & 33.05              & 0.9620              & 0.0405              \\
  & SAM-3D-Body (MHR)  & SMPL-X & \underline{35.04}  & \underline{0.9724}  & \underline{0.0333}  \\
  & SAM-3D-Body (MHR)  & MHR    & \textbf{37.56}     & \textbf{0.9787}     & \textbf{0.0267}     \\
\bottomrule
\end{tabular}%
}
\end{table}

\subsection{Disentangling Pose Quality and Mesh Capacity}
\begin{figure}[htbp]
    \centering
    \includegraphics[width=0.5\textwidth]{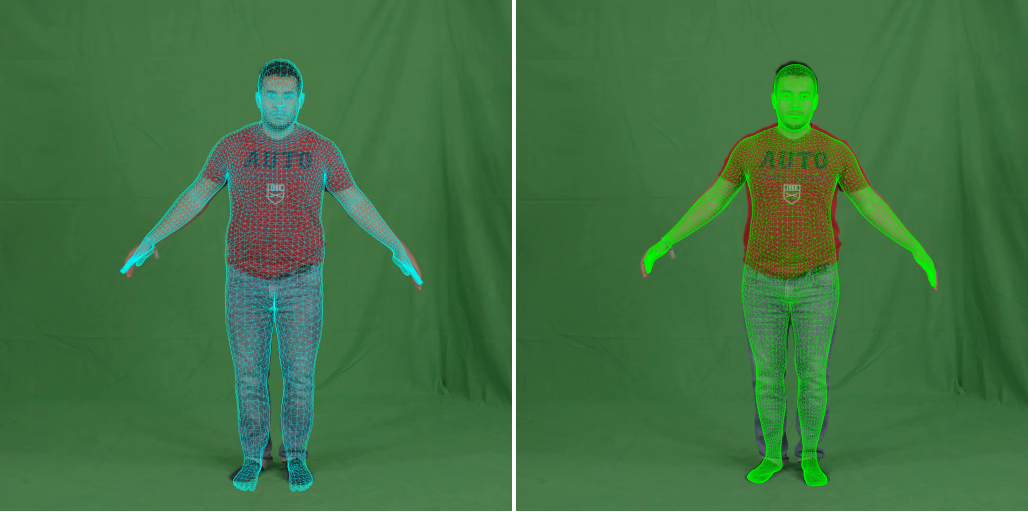}
    \caption{Render of pose provided by Anim-NeRF (left) compared to the pose estimated with SAM-3D-Body (right).}
    \label{fig:smpl_vs_mhr}
\end{figure}

Our main results conflate two factors: SAM-3D-Body may produce more accurate poses than the Anim-NeRF fits used by prior methods, and MHR's mesh may better support Gaussian attachment than SMPL-X.
Figure \ref{fig:smpl_vs_mhr} illustrates the pose estimation gap. 
The SAM-3D-Body fit (right) captures hand orientation, head shape, and foot placement more faithfully than the Anim-NeRF fit (left), and its mesh contours follow the underlying human shape rather than the clothing silhouette.
Table \ref{tab:pose_quality} isolates each factor on the two PeopleSnapshot subjects for which Anim-NeRF's optimized SMPL parameters are publicly available via links provided by the Anim-NeRF team.
Averaged across both subjects, upgrading pose estimation alone accounts for +2.50 PSNR, +0.0123 SSIM, and $-$0.0070 LPIPS, while upgrading the rendering mesh contributes a further +2.41 PSNR, +0.0080 SSIM, and $-$0.0080 LPIPS, confirming that both factors provide independent, roughly comparable gains.

\section{Conclusion}
\label{sec:Conclusion}

Body model quality is an underexplored axis of improvement in the Gaussian avatar literature. 
Our results demonstrate that MHR's representational advantages (twist joints, non-linear pose correctives, and higher mesh density) are sufficient to match or exceed methods that compensate for SMPL's failure modes with learned deformation modules, rectification networks, and pose-dependent MLPs.
Our controlled ablation, retaining SAM-3D-Body poses while translating to SMPL-X, confirms that representational capacity contributes independently of pose estimation quality as identical poses consistently yield better results when utilizing MHR.

This finding suggests the field's architectural innovations, while effective, have been partially compensating for upstream body model deficiencies. 
Upgrading the body model may be a simpler and more direct path to quality gains than adding downstream complexity.

Our study has several limitations. 
Our pipeline deliberately disables all learned corrections to isolate the body model's contribution and consequently falls short of leading methods on SSIM and LPIPS.
The remaining perceptual gap indicates these corrections still provide value, particularly for pose-dependent texture detail.
We compare only SMPL-family models and MHR; other body models such as STAR~\cite{STAR} and GHUM~\cite{GHUM} may offer different tradeoffs.
MHR's ecosystem is also far less mature than SMPL's, though SAM-3D-Body provides a practical estimation path that requires no custom fitting pipeline. 

Three directions follow naturally from this work. 
First, combining MHR with lightweight pose-dependent corrections should close the remaining perceptual gap while preserving the geometric foundation that MHR provides. 
Second, ablating individual MHR features would reveal which aspects of body model design matter most for downstream avatar quality, with practical implications for future body model development.
Finally, updating existing methods with improved poses and MHR could reveal how much of their learned deformation capacity becomes redundant when the upstream body model already handles twist corrections and fine-grained surface detail, potentially enabling simpler and faster architectures without sacrificing quality.
{
    \small
    \bibliographystyle{ieeenat_fullname}
    \bibliography{main}
}


\end{document}